\documentclass[letterpaper, 10 pt, conference]{ieeeconf}  

\usepackage[utf8]{inputenc}
\usepackage{graphicx}
\usepackage{amsmath}   
\usepackage{amsfonts}  
\usepackage{amssymb}   
\usepackage{bm}        

\IEEEoverridecommandlockouts                              
\title{\LARGE \bf
Unsupervised Learning of Inter-Object Relationships via Group Homomorphism
}

\author{Kyotaro Ushida$^{1}$, Takayuki Komatsu$^{1}$, Yoshiyuki Ohmura$^{1}$ and Yasuo Kuniyoshi$^{1}$
\thanks{The authors are with the Mechano-Informatics Department, Graduate School of Information Science and Technology, The University of Tokyo, Tokyo 113-8656, Japan (e-mail: k-ushida@isi.imi.i.u-tokyo.ac.jp; komatsu@isi.imi.i.u-tokyo.ac.jp; ohmura@isi.imi.i.u-tokyo.ac.jp; kuniyosh@isi.imi.i.u-tokyo.ac.jp).}
}

\begin{document}

\maketitle
\thispagestyle{empty}
\pagestyle{empty}

\begin{abstract}

While current deep learning models achieve high performance by learning statistical correlations from vast datasets,which stands in stark contrast to human learning.
They lack the flexibility of humans—particularly preverbal infants—to autonomously acquire the underlying structure of the world from limited experience and adapt to novel situations. 
In this study, we propose an unsupervised representation learning method based on a hierarchical relationship in group operations, rather than statistical independence, aiming to build a computational model of the cognitive development of infants. 
The proposed model features an integrated architecture that simultaneously performs object segmentation and the extraction of motion laws from dynamic image sequences. By introducing the Homomorphism from algebra as a structural constraint within a neural network, the model structurally separates pixel-level changes into meaningful, decomposed transformation components, such as translation and deformation. 
Using interaction scenes (chasing and evading tasks) based on developmental science findings, we experimentally demonstrate that the model can segment multiple objects into individual slots without any ground-truth labels. Furthermore, we confirmed that relative movements between objects, such as approaching or receding, are accurately mapped and structured into a one-dimensional additive latent space. These results suggest that by introducing algebraic geometric constraints rather than relying solely on statistical correlation learning, physically interpretable "disentangled representations" can be acquired. This study contributes to the understanding of the process by which infants internalize environmental laws as structures and provides a new perspective for constructing artificial systems with developmental intelligence.

\end{abstract}

\section{INTRODUCTION}

In recent years, advancements in deep learning have enabled artificial intelligence (AI) to achieve superhuman performance in specific tasks such as natural language processing and image recognition.
However, these technologies remain heavily dependent on enormous computational resources and data, facing significant challenges regarding vulnerability to novel situations outside their training distribution. 
This fragility stems from the fact that current AI processes data primarily as statistical correlations rather than as underlying structures.

In contrast, human intelligence—particularly in preverbal infants—possesses the ability to autonomously structure and understand the world through limited, continuous experiences in early life. Rather than merely memorizing patterns, infants build the foundation of versatile intelligence by "structuring" their understanding of object permanence and the laws of motion, allowing them to adapt flexibly to unknown scenarios. Therefore, for AI to acquire truly flexible intelligence, a "constructive" approach \cite{piaget1963origins} that captures the environment as a structure, transcending simple statistical learning, is indispensable.

In the field of representation learning, obtaining the latent structures of data as "disentangled representations" is a primary objective. While conventional methods have predominantly aimed for disentanglement through statistical learning\cite{kingma2013auto,goodfellow2014generative}, they face inherent limitations in representing complex structural relationships, such as logical dependencies or non-commutative concepts. To overcome these limitations,
 this study assumes group structures and applies group theory to representation learning.
Humans recognize dynamic visual information by structuring elements such as color, shape, and motion as attributes. When humans recognize an object, they understand that even if its position changes, the inherent properties of the object remain unchanged. As a mathematical framework for handling this "relationship between transformation and conservation," there are symmetries in group theory: invariance and equivariance.
Invariance refers to the property where an object's characteristics are maintained even when certain operations or transformations are applied, while equivariance means that the representation of an object changes in a manner corresponding to the applied transformation. Learning such "disentangled representations," where elements are appropriately separated, is the key for AI to acquire intelligence that possesses human-like flexibility and high generalization performance without relying heavily on statistical learning.

Humans are thought to bring information into consciousness by extracting semantic structures from the environment, and this process of structuring is strongly related to consciousness.
Ohmura et al. have proposed the Dual-Laws Model (DLM) \cite{ohmura2024mentalcausationphysicalworld,ohmura2025consciousnessexplainphysicalphenomena,ohmura2026mathematicalformalizationselfdeterminingagency} as a new theory of artificial consciousness and cognition. Based on the assumption that mental processes supervene on physical systems (the lower base level), the DLM attempts to formalize consciousness as a hierarchical system where "micro-physical laws" and "macro-dynamical laws" coexist. The core of this theory is that the macro level, while dependent on the dynamics of the micro level, possesses its own independent dynamics and can impose inter-level causality on micro-states. Ohmura et al. suggest that "algebraic structures" can function as a macro-law constraint, potentially controlling micro-neural activity through cognitive mechanisms such as attention. 
We believe that this DLM framework is highly effective for constructing artificial cognition capable of autonomously structuring input data from the environment.
While this theory remains empirically under-validated, this paper demonstrates the efficacy of group homomorphisms as algebraic constraints within its framework.

By constraining continuous pixel-level changes (micro) with algebraic symmetries and laws (macro), it should be possible to build robust internal models of the world that go beyond ad-hoc statistical learning.

In this study, we introduce "Group Homomorphism" as a specific implementation of these macro-level constraints. By integrating this constraint with unsupervised object segmentation, we propose a representation learning method that simultaneously recognizes individual objects in image sequences and extracts the algebraic relationships between their motions.

The ultimate goal of this research is to constructively model the process of cognitive development in preverbal infants. As a fundamental requirement for understanding "Agency" in oneself and others, this study addresses three essential tasks: the separation of transformations (such as motion), object segmentation, and the classification of relationships between multiple objects. Specifically, using image sequences as input, we propose an unsupervised representation learning method consisting of the following three steps:

\begin{enumerate}
\item \textbf{Object Segmentation:} Generating segmentation masks to isolate individual objects from image sequences.
\item \textbf{Separation of Object Motion:} Using group homomorphism constraints to separate the motion of isolated objects into distinct components, such as translation and deformation.
\item \textbf{Extraction of Multi-Object Interactions:} Relativizing the motion of each object from the perspective of others to extract underlying interactions.
\end{enumerate}

Through these steps, we propose a representation learning method to structurally capture the motion of multiple objects within image sequences.

\section{RELATED WORKS}

The evolution of representation learning has transitioned from capturing linear statistical features, such as PCA and ICA, to modeling complex high-dimensional manifolds using deep generative models like VAEs \cite{kingma2013auto, higgins2017beta} and GANs \cite{goodfellow2014generative}. While powerful, these frameworks primarily rely on statistical independence and often struggle to capture hierarchical or non-independent structural relationships.

To address this, object-centric learning, exemplified by Slot Attention \cite{locatello2020objectcentriclearningslotattention} and SAVi \cite{kipf2022conditionalobjectcentriclearningvideo}, has emerged to decompose scenes into individual "slots." However, while these models enable object tracking through temporal continuity, they lack an explicit mechanism to semantically disentangle object motions or formalize multi-object interactions.

Furthermore, research such as PLATO \cite{piloto2022intuitive} has incorporated cognitive science insights to demonstrate that intuitive physics—concepts like object permanence and solidity—can be learned through the minimization of prediction error. However, many of these approaches either require supervised information for segmentation or represent physical laws only "implicitly" within the neural network weights. To achieve human-like abstract conceptual understanding, it is essential to take an approach that "explicitly" formalizes physical phenomena as algebraic structures, such as group theory.

Our research group has developed methods to structurally separate transformations based on algebraic structures. Early work utilized the commutativity of transformations to disentangle attributes such as color and shape \cite{ohmura2023algebraictheorydiscriminatequalia}. More recent studies have introduced constraints based on group homomorphisms, enabling the unsupervised separation of non-commutative transformations, such as simultaneous translation and rotation \cite{nishitsunoi2025learning}:
\begin{equation}
\rho(g_1 \circ g_2) = \rho(g_1) \cdot \rho(g_2)
\end{equation}
The present study inherits this lineage and extends the scope from the "description of transformations for a single object" to a new stage: "representing the interactions between multiple objects."

\section{METHOD}

In this section, we propose a novel constructive approach based on group homomorphisms to model the cognitive development of preverbal infants. 
Our goal is to enable an agent to autonomously structure environmental transformations into an operable algebraic framework (group structure) through sensorimotor experience, moving beyond the implicit physical modeling found in prior works. 
The proposed model consists of three integrated components: (1) unsupervised object segmentation, (2) transformation separation via group homomorphisms, and (3) relational representation learning between multiple objects.

\subsection{Transformation Separation via Group Homomorphism}
\label{sec:method:homomorphism}
To disentangle meaningful components such as translation or deformation from image variations, we introduce a method to map the structure of image changes into a lower-dimensional representation space as a group homomorphism, following Komatsu et al. \cite{komatsu2026transformation}.

\subsubsection{Formulation and Extraction of Transformations}
For a sequence of image pairs $(x, x')$, we estimate the element $g \in G$ that induces the observed change. 
Specifically, following the previous work \cite{komatsu2026transformation}, we construct an encoder $\Phi$ that computes the transformation parameter $g$ from the estimated transformation field $\mathcal{T}_g$. 
We then apply a mapping $\rho$ (implemented as a neural network) to $g$ to obtain a representation $h = \rho(g)$ in the low-dimensional latent space $H$.

\subsection{loss functions}
To ensure that the representation space $H$ preserves the algebraic group structure and extracts meaningful information, we minimize the following two loss functions:

Homomorphism Loss:

This constraint ensures that the binary operation $\cdot$ in the representation space is consistent with the composition $\circ$ of the original transformations:
$$\mathcal{L}_{\text{homo}} = \| \rho(g_1 \circ g_2) - (\rho(g_1) \cdot \rho(g_2)) \|^2 \quad $$

Variance Loss:

To prevent "representation collapse"—where all representations converge to the identity element—this loss forces the latent space to maintain a diverse set of informative features:
$$\mathcal{L}_{\text{var}} = \| \text{ReLU}(m - \sigma(\mathbf{h})) \|^2_2 \quad $$

\begin{figure}
    \centering
    \includegraphics[width=\linewidth]{img/model_homo.png}
    \caption{Conceptual diagram of motion separation based on group homomorphism.
    The transformation between an image pair $(x, x')$ is conceptualized as a transformation field $T_g$, parameterized by an element $g$ of the transformation group $(G, \circ)$. The model extracts this element $g$ from the identified $T_g$ using an encoder $\Phi$, while a decoder $\Psi$ acts in the reverse direction. The core of our method is the introduction of a homomorphism $\rho$ from group $G$ to a representation group $H$. This algebraic constraint enables the model to selectively extract components with specific algebraic properties into $H$, while filtering other transformations into the kernel of the mapping.}
    \label{fig:homo-model}
\end{figure}

\subsection{Unsupervised Object Segmentation} \label{sec:method:segmentation}

To extract individual objects without labels, we employ a U-Net architecture \cite{ronneberger2015u} that outputs $K$ attention masks $\{m_o\}_{o=1}^N$. We define the masked object image as $x_o = x \odot m_o$. 
To ensure the masks capture meaningful "object-like" entities, we minimize the following unsupervised loss:
\begin{equation}
    \mathcal{L}_{\text{seg}} = \lambda_{\text{div}} \mathcal{L}_{\text{div}} + \lambda_{\text{bin}} \mathcal{L}_{\text{bin}} + \lambda_{\text{area}} \mathcal{L}_{\text{area}}
\end{equation}
where $\mathcal{L}_{\text{div}}$ enforces spatial exclusivity between masks (diversity), $\mathcal{L}_{\text{bin}}$ encourages pixel values toward 0 or 1 (binarization), and $\mathcal{L}_{\text{area}}$ penalizes excessively large masks.
The segmentation module is integrated with the transformation module by minimizing a prediction reconstruction loss:
\begin{equation}
    \mathcal{L}_{\text{pred recon}} = \left| x^{t+1} - \sum_{o=1}^N (\mathcal{T}_{\bm{g}_o} \cdot x_o^t) \right|^2_2
\end{equation}
This forces the model to segment objects correctly so that their individual motions ($\bm{g}_o$) can accurately reconstruct the next frame.

\begin{figure*}[htb]
    \centering
    \includegraphics[width=\linewidth]{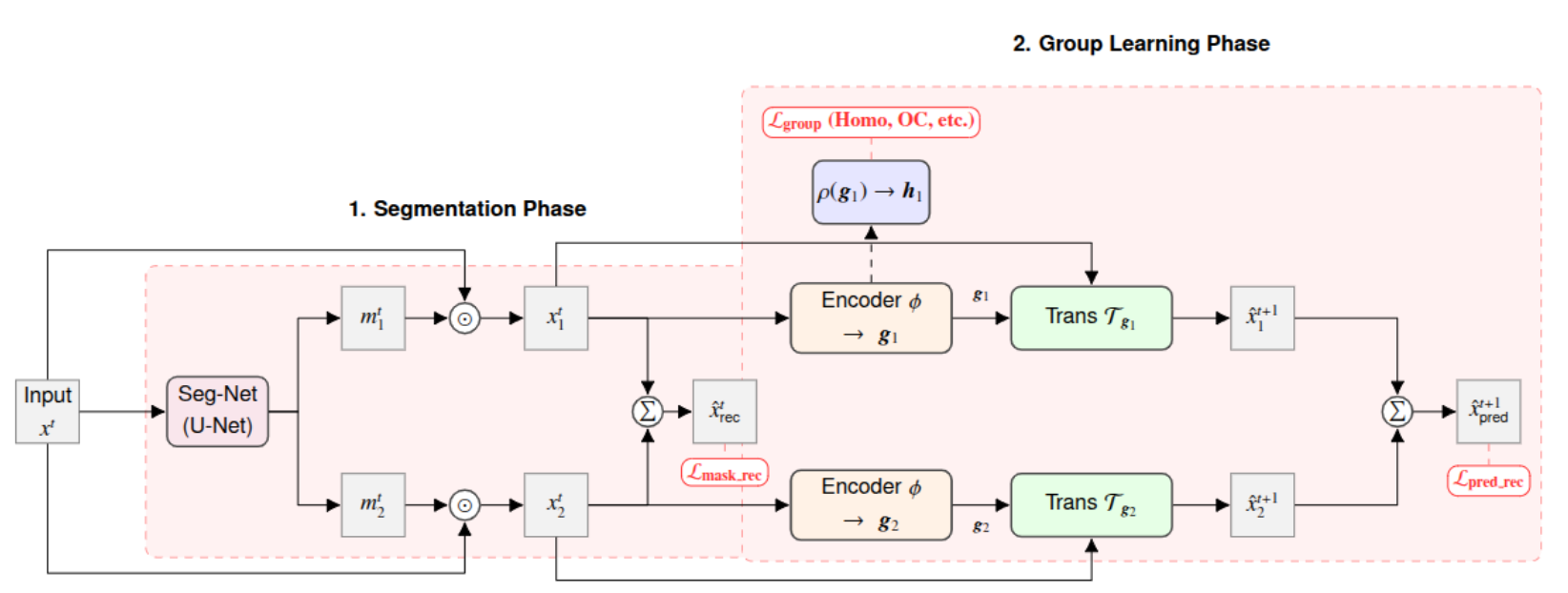}
    \caption{\ref{sec:method:segmentation}:Segmentation architecture. 
    Segmentation Phase: A Seg-Net (U-Net) generates multiple attention masks $m_k^t$ from the input image $x^t$ to extract individual objects $x_k^t$.Group Learning Phase: For each extracted object, an encoder $\phi$ estimates a transformation group element $g_k$. The group homomorphism constraint $\rho(g) \to h$ is applied to this latent space to ensure structured representation learning.Finally, the predicted next frame $\hat{x}^{t+1}_{\text{pred}}$ is reconstructed by applying the estimated transformations $\mathcal{T}_{g_k}$ to each object. The entire system is trained end-to-end by minimizing the prediction reconstruction loss $\mathcal{L}_{\text{pred\_rec}}$}
    \label{fig:seg-model}
\end{figure*}

\subsection{Representation of Inter-Object Relationships}
\label{sec:method:relation}

\subsubsection{Formulation of Relative Motion}
To understand the interaction between objects $O_1$ and $O_2$, we define a relative transformation parameter $\bm{g}_{\text{rel}} = \bm{g}_1^{-1} \circ \bm{g}_2$ in the parameter space $G$. 
To capture the interactions between objects in a scene, this study shifts the description of motion from an absolute coordinate system to a relative coordinate system centered on a specific object.
 Given two objects $x_1$ and $x_2$ with their respective global transformation parameters $g_1, g_2 \in G$, the generative process of the entire image can be described as follows:
 $$x_{total} = T_{g_1} \cdot x_1 + T_{g_2} \cdot x_2$$
 This process can be mathematically reformulated to represent the scene from the perspective of either object. 
 For instance, when $x_1$ is taken as the reference frame (viewpoint), the equation becomes:$$x_{total} = T_{g_1} \cdot (x_1 + (T_{g_1^{-1}} \circ T_{g_2}) \cdot x_2)$$
 Conversely, if $x_2$ is chosen as the reference, the relationship is symmetrically described as:$$x_{total} = T_{g_2} \cdot ((T_{g_2^{-1}} \circ T_{g_1}) \cdot x_1 + x_2)$$
 In these derivations, the expressions inside the brackets represent the relative motion of one object as seen from the frame where the other is stationary. 
 This allows the model to effectively fix its "attention" on a specific object and extract relative relationships grounded in that object’s frame of reference. 
 The relative transformation parameter $g_{rel}$ is thus defined based on the chosen subject-object relationship:$$g_{2|1} = g_1^{-1} \circ g_2 \quad \text{or} \quad g_{1|2} = g_2^{-1} \circ g_1$$
 Our model maps these relative parameters into the representation space $H$ via the homomorphism $\rho$, yielding the latent relational vector $h_{rel} = \rho(g_{rel})$. 
 This operation algebraically formalizes the "viewpoint" shift, where the choice of the reference object determines the coordinate system for the entire interaction.

\subsubsection{Projection to Interpretable scalar}
We map this relative parameter into the representation space $H$ using the homomorphism $\rho$ defined in Section \ref{sec:method:homomorphism} (following \cite{komatsu2026transformation}), yielding the latent relational vector $h_{\text{rel}} = \rho(g_{2|1})$.
To further extract interpretable physical quantities, we learn a secondary projection $P: H \to S$ from the high-dimensional space $H$ to a low-dimensional space $S$ (e.g., a scalar $\mathbb{R}$). 

By imposing an additional homomorphism constraint $P(h_1 \cdot h_2) = P(h_1) \oplus P(h_2)$, where $\oplus$ denotes simple addition, the projected value $s = P(h_{\text{rel}})$ is encouraged to correspond to additive physical quantities such as relative velocity or angular change. 
Similar to the previous sections, we apply a variance loss $\mathcal{L}_{\text{var}}$ to prevent trivial solutions.

\section{EXPERIMENTS}

In this section, we evaluate the effectiveness of the proposed method—unsupervised multi-object segmentation and relational representation acquisition based on group homomorphisms. 
Using dynamic interaction scenes that model the cognitive development of preverbal infants, we assess whether the individuality of objects and the structure of their interactions can be autonomously acquired through unsupervised learning.

\subsection{Experimental Setup}

\subsubsection{Dataset Generation}
To learn the perception of interactions, we constructed a custom image sequence dataset using Pygame, mimicking the experimental setup of Kanakogi et al. \cite{kanakogi2022third}. 

\begin{figure}[ht]
    \centering
    \includegraphics[width=0.5\linewidth]{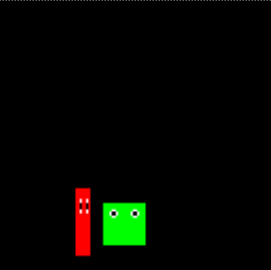}
    \caption{Generated image sequences of interaction scenes. The dataset consists of two agents: a Chaser (rigid body) and an Evader (soft body) and they have "eyes".}
\end{figure}

\begin{figure}
    \centering
    \includegraphics[width=\linewidth]{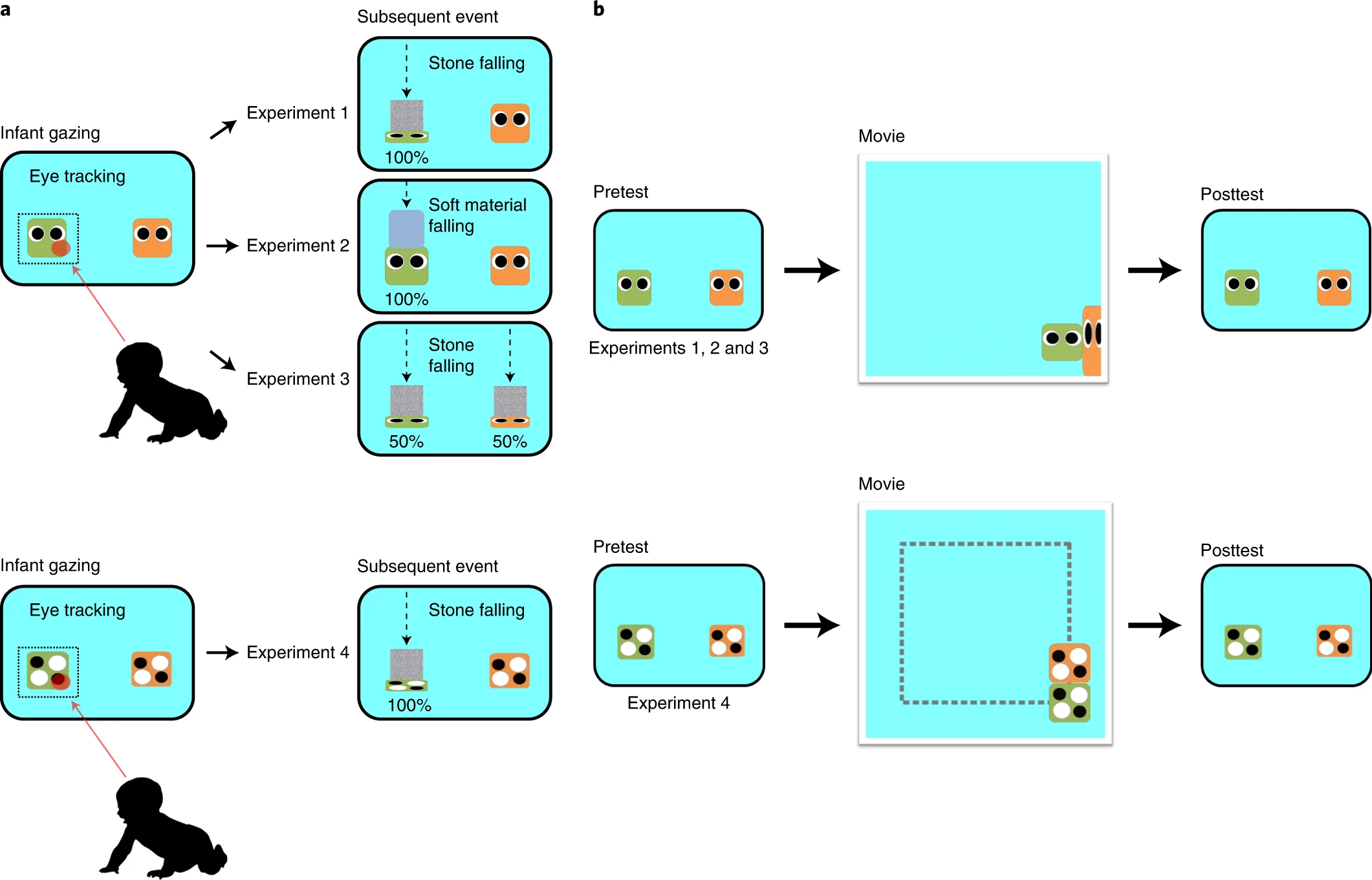}
    \caption{Experimental setup from the prior study \cite{kanakogi2022third}, which served as the basis for our dataset generation.
    The conceptual diagram illustrates an experiment using videos where agents with "eyes" are depicted, with one agent chasing and colliding with (attacking) another. This study demonstrated that preverbal infants, after observing these videos, make choices that effectively "punish" the agent exhibiting aggressive behavior. In our research, we utilized a dataset consisting of randomly generated image sequences that simulate this experimental paradigm.}
    \label{fig:input_img_sequences}
\end{figure}

Each agent moves according to physical calculations with the following characteristics:

\begin{itemize}
    \item \textbf{Chaser:} High mobility aimed at the target. It behaves as a rigid body and does not deform upon collision.
    \item \textbf{Evader:} Generates acceleration to move away from the chaser. While its mobility is lower, it behaves as a soft body that deforms upon collision based on physical parameters.
\end{itemize}

During collisions, we implemented momentum conservation laws, a "frozen state" to enhance visibility of the event, and "Squash and Stretch" animations that approximately satisfy area conservation (Fig. \ref{fig:input_img_sequences}).

\subsubsection{Model Configuration}
The model integrates the segmentation module and the group homomorphic transformation separation module. We adopted a curriculum learning strategy, introducing the homomorphism loss $\mathcal{L}_{H}$ after 30,000 steps. 
Optimization was performed end-to-end to minimize the sum of the reconstruction error and algebraic constraints.

\subsection{Experiment 1: Unsupervised Multi-Object Segmentation}
\label{subsec:exp1}
In this experiment, we evaluate the effectiveness of the proposed unsupervised segmentation module described in Section \ref{sec:method:segmentation}. Using the dataset generated in the previous section, we demonstrate that the model can successfully segment and track multiple moving objects without any ground-truth labels.

\subsubsection{Results}
\begin{figure}[ht]
    \centering
    \includegraphics[width=0.8\linewidth]{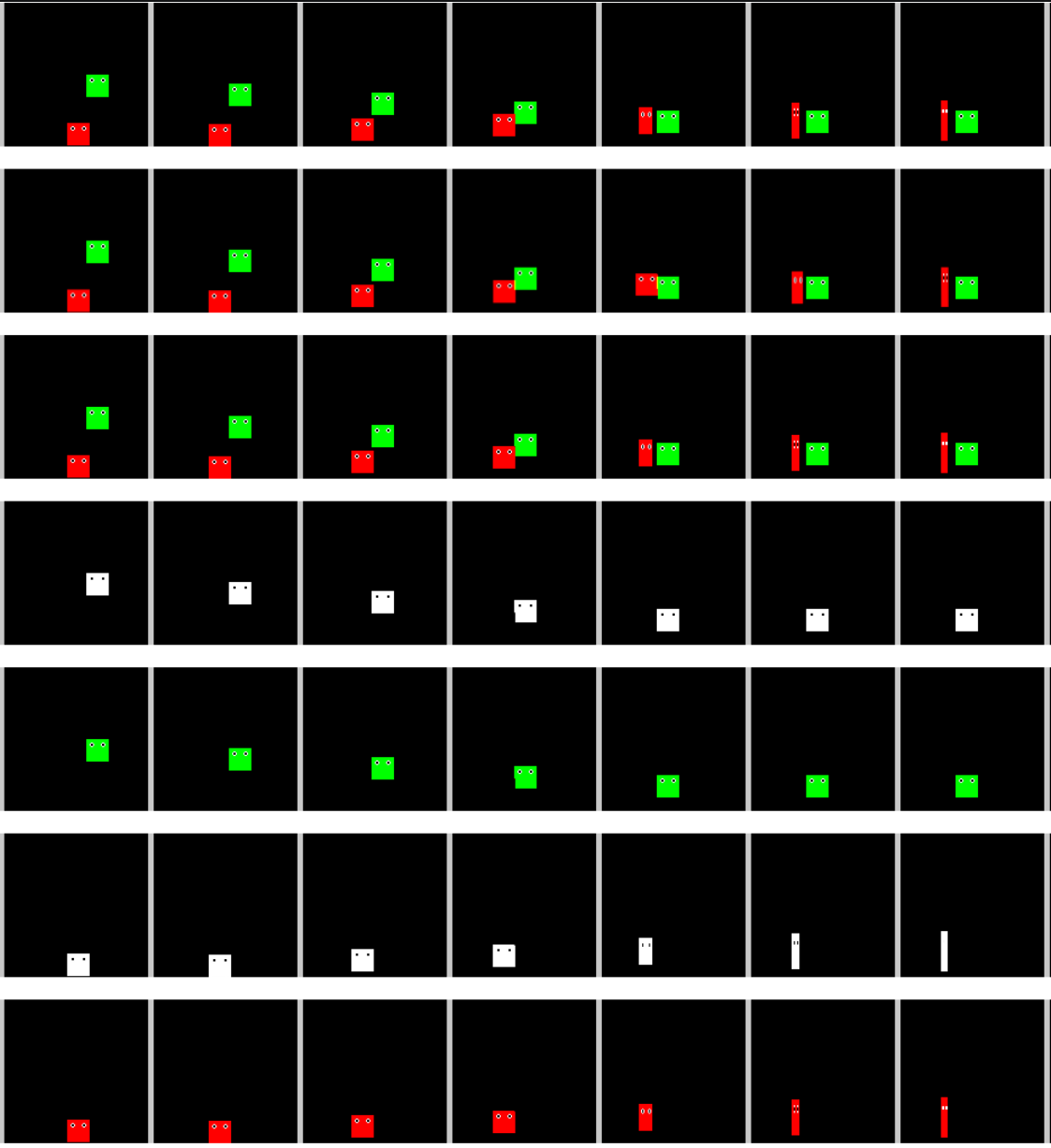}
    \caption{Segmentation results. It successfully separated the Chaser and Evader into different slots without any labels.}
    \label{fig:exp1}
\end{figure}

The experimental results demonstrated that, although dependent on the initial random seed, the proposed model succeeded in appropriately segmenting individual objects without supervision.
 In the early stages of learning (around Step 40,000), "degeneracy" occurred where multiple objects were covered by a single mask. 
 However, as learning progressed (Step 150,000), the Chaser and Evader were assigned to distinct slots, obtaining clear segmentation masks separated from the background (Fig. \ref{fig:exp1}).

\subsubsection{Discussion}
The primary factor enabling autonomous separation without pixel labels is the task setting of "transformation prediction." Since the Chaser and Evader follow different motion laws (velocity and acceleration vectors), attempting to explain both with a single transformation parameter increases reconstruction error. 
To minimize this error, the model acquired an inductive bias to treat regions governed by the same group action (motion law) as a single object. 
Furthermore, the model's flexibility allowed it to track the Evader's non-rigid deformations during collisions without breaking object identity.

\subsection{Experiment 2: Representation of Inter-Object Relationships}
\label{subsec:exp2}
Building on the segmentation results from Section \ref{sec:method:segmentation}, this experiment demonstrates that the relational module described in Section \ref{sec:method:relation} successfully constructs an interpretable scalar $s$ that represents the concepts of "approaching" and "receding."

\subsubsection{Results}
\begin{figure}[ht]
    \centering
    \includegraphics[width=\linewidth]{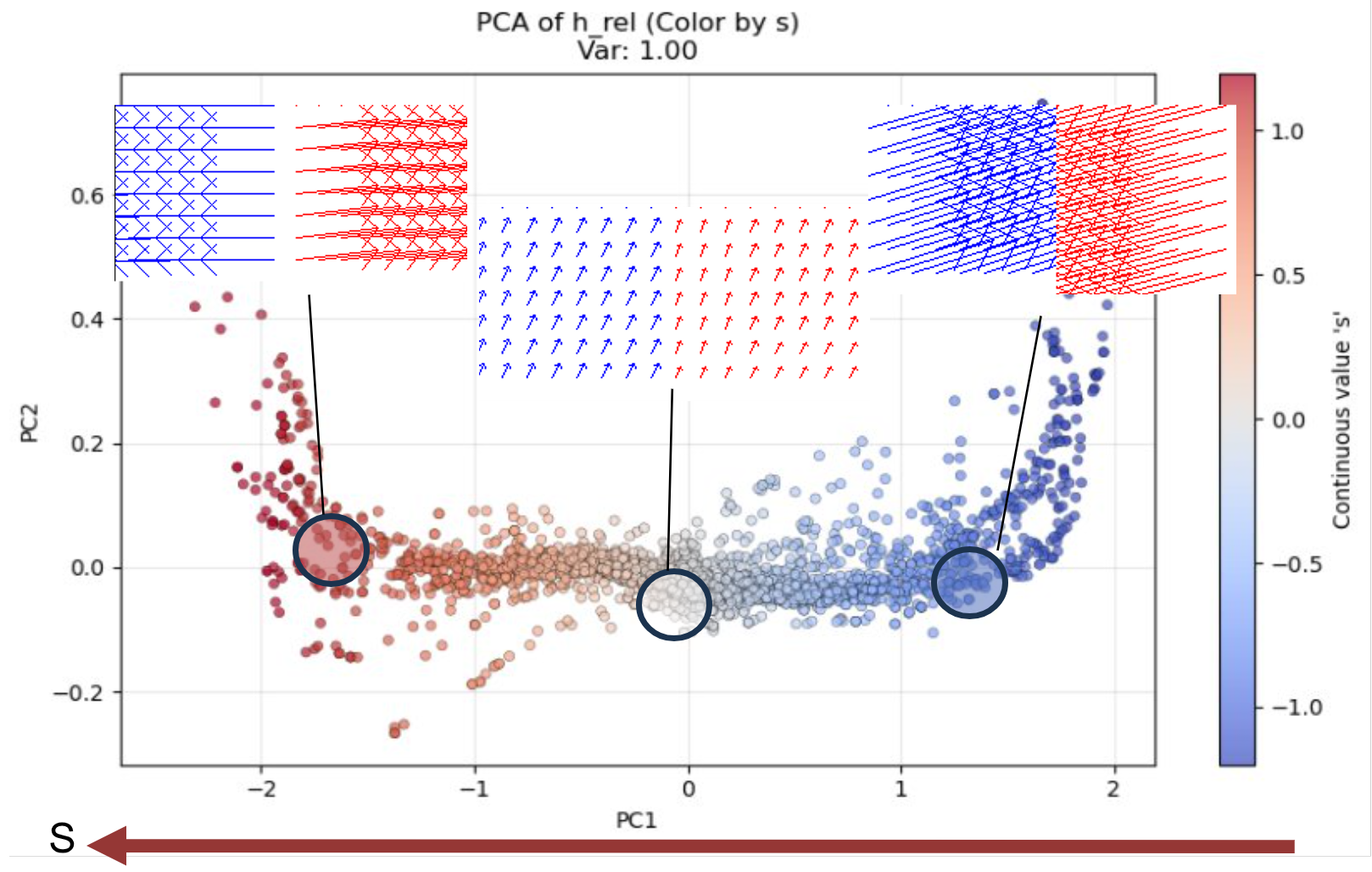}
    \caption{Visualization of inter-object relationship representation acquired via group homomorphism.
The relative transformation $h_{\text{rel}}$ in the representation group $H$ is projected onto a 2D plane using Principal Component Analysis (PCA). The distribution is colored by the scalar value $s$, obtained through an additional homomorphism mapping from $H$ to the scalar addition group $S$. Larger positive values of $s$ (red) correspond to outward motions where the two objects recede from each other, while smaller negative values (blue) correspond to inward motions where they approach each other. This demonstrates that the algebraic constraint forces the model to organize complex transformations into a structured scalar axis representing relative displacement.}
    \label{fig:exp2}
\end{figure}

Fixing the encoder parameters from Experiment 1, we performed additional training of a module to extract the relative relationship $s \in S$ (scalar space) between objects. We visualized the acquired relationship representation $s$ by performing Principal Component Analysis (PCA) on the relative transformation representation $h_{rel} = h_A^{-1} \cdot h_B$ and coloring the distribution by the value of $s$ (Fig. \ref{fig:exp2}).

The results demonstrate that the distribution of $h_{rel}$ formed a U-shaped structure on a 2D plane. Crucially, the acquired $s$ values changed smoothly and monotonically along the first principal component (PC1). Specifically, large values of $s$ corresponded to agents receding from each other (Fig. \ref{fig:exp2}), small values corresponded to approaching (Fig. \ref{fig:exp2}), and values near zero corresponded to parallel movement where the relative distance remained constant.

\subsubsection{Discussion}
These results suggest that the algebraic constraint of group homomorphism ($f(a \cdot b) = f(a) + f(b)$) functions as a powerful structural bias in representation learning. The model is forced to map complex non-linear changes in data space onto a "linear axis where operations are defined (e.g., distance or displacement)." This indicates that algebraic constraints can contribute significantly to the disentanglement of representations.

However, certain limitations were identified. The $s$ extracted here primarily describes geometric changes in relative position (symmetrical relationships). To describe social relationships such as "who is chasing whom," the current group-theoretic framework—especially one assuming commutative operations—may lack the constraints necessary to capture the asymmetry of Agency. Additionally, since the representation relies solely on motion (transformations), it remains difficult to identify specific relationships without state information like absolute position. Integrating algebraic motion structures with object state representations will be a key challenge for achieving higher-level interaction cognition.

\section{CONCLUSIONS}

This study proposed an unsupervised representation learning framework based on group homomorphisms for the computational modeling of preverbal cognitive development. 
By integrating object segmentation with motion law extraction, our model achieves autonomous structuring of dynamics from raw image sequences through algebraic constraints.

The primary contribution of this research is the demonstration that group-theoretic constraints allow an agent to autonomously structure "inter-object relationships" without any ground-truth labels. While prior works \cite{nishitsunoi2025learning,komatsu2025featurebasedliegrouptransformer}, successfully extracted the dynamics of a single transformation, our model extends this algebraic framework to multi-object scenes.
Specifically, we confirmed that the model not only segments multiple moving objects based on their distinct motion laws but also maps their relative interactions (e.g., approaching or receding) into a structured, one-dimensional additive latent space.
This proves that physically interpretable "disentangled representations" can be acquired through geometric homomorphisms rather than relying solely on conventional statistical correlations.

As a future challenge, while this study focused on disentangling temporal motions, integrating state information—such as object positions and orientations—is essential for more precise relational descriptions. Specifically, extending the current analysis of "temporal changes" to "inter-object differences" at a single point in time represents a crucial next step.

A fundamental limitation remains: descriptions based on pixel-level transformation fields struggle with objects of diverse textures and shapes, yet an over-reliance on deep features risks creating representations tied to specific appearances. Therefore, a key challenge is to develop feature-independent methods that capture underlying geometric and algebraic relationships, regardless of visual attributes like color or texture.

In conclusion, these findings contribute to understanding how infants autonomously acquire the physical and geometric laws of the external world, serving as a significant step toward the construction of artificial systems endowed with developmental intelligence.




\section*{ACKNOWLEDGMENT}
This work was supported by JSPS KAKENHI Grant Number JP[25H00448].



\begin{normalsize}
\bibliographystyle{bib/IEEEtranBST/IEEEtran} 
\bibliography{bib/IEEEtranBST/IEEEabrv,bib/liter}
\end{normalsize}

\end{document}